\documentclass[conference,10pt]{IEEEtran}
\IEEEoverridecommandlockouts
\usepackage{cite}
\usepackage{amsmath,amssymb,amsfonts}
\usepackage{algorithmic}
\usepackage{graphicx}
\usepackage{textcomp}
\usepackage{xcolor}
\usepackage{booktabs}
\usepackage{multirow}
\usepackage{pifont}
\usepackage{xcolor}
\usepackage{listings}
\usepackage{algorithm}
\usepackage{algorithmic}
\usepackage{authblk}
\usepackage{mweights}
\usepackage{newtxtext}
\usepackage{array,tabularx}
\usepackage[pagebackref,hidelinks]{hyperref}

\lstdefinestyle{pythonstyle}{
        language=Python,
        basicstyle=\ttfamily\bfseries,
        keywordstyle=\color{blue}\bfseries,
        commentstyle=\color{green!50!black}\itshape,
        stringstyle=\color{orange},
        tabsize=4,
        captionpos=b
}

\usepackage{xcolor}
\definecolor{mycitecolor}{RGB}{46, 99, 158} 
\newcolumntype{C}{>{\centering\arraybackslash}X}
\newcolumntype{R}{>{\raggedleft\arraybackslash}X}
\newcolumntype{L}{>{\raggedright\arraybackslash}X}
\hypersetup{
    colorlinks=true,    
    linkcolor=black,  
    citecolor=black,  
    urlcolor=black    
}

\title{HSR-KAN: Efficient Hyperspectral Image Super-Resolution via Kolmogorov-Arnold Networks}
\author[1, 2]{Baisong Li}
\author[1, 2*]{Xingwang Wang \thanks{*Corresponding Author}}
\author[1, 2]{Haixiao Xu}
\affil[1]{College of Computer Science and Technology, Jilin University}
\affil[2]{Key Laboratory of Symbolic Computation and Krowledge Engineering \cr
of Ministry of Education, Jilin University }
\affil[ ]{ \tt{lbs23@mails.jlu.edu.cn, \{xww,haixiao\}@jlu.edu.cn} }

\begin{document}
\maketitle
\begin{abstract}
Hyperspectral images (HSIs) have great potential in various visual tasks due to their rich spectral information. However, obtaining high-resolution hyperspectral images remains challenging due to limitations of physical imaging. Inspired by Kolmogorov-Arnold Networks (KANs), we propose an efficient HSI super-resolution (HSI-SR) model to fuse a low-resolution HSI (LR-HSI) and a high-resolution multispectral image (HR-MSI), yielding a high-resolution HSI (HR-HSI). To achieve the effective integration of spatial information from HR-MSI, we design a fusion module based on KANs, called KAN-Fusion. Further inspired by the channel attention mechanism, we design a spectral channel attention module called KAN Channel Attention Block (KAN-CAB) for post-fusion feature extraction. As a channel attention module integrated with KANs, KAN-CAB not only enhances the fine-grained adjustment ability of deep networks, enabling networks to accurately simulate details of spectral sequences and spatial textures, but also effectively avoid Curse of Dimensionality. Extensive experiments show that, compared to current state-of-the-art HSI-SR methods, proposed HSR-KAN achieves the best performance in terms of both qualitative and quantitative assessments. Our code is available at: https://github.com/Baisonm-Li/HSR-KAN.
\end{abstract}


\section{Introduction}
Hyperspectral images (HSIs) contain rich spectral bands, providing more precise and detailed spectral information compared to traditional RGB images. Leveraging this advantage, HSIs play a crucial role in various fields of computer vision and are widely applied in tasks such as image classification~\cite{class_02, class_03}, scene segmentation~\cite{seg_01,seg_02}, and object tracking~\cite{Tracking_01, OBJ_02}. However, due to the limitations of physical imaging, obtaining high-resolution hyperspectral images remains a significant challenge. This difficulty primarily arises from constraints in sensor hardware, the complexity of high-dimensional data acquisition, and the inherent trade-off between spatial and spectral resolution. Consequently, hyperspectral image super-resolution (HSI-SR) has emerged as a key research direction in this field.

In recent years, numerous HSI-SR methods have been proposed, mainly categorized into four types: Bayesian inference-based~\cite{bayesian_01,bayesian_02}, matrix decomposition-based~\cite{matrix_01}, tensor-based~\cite{Tensor_01}, and deep learning (DL)-based methods~\cite{hsr-net, fusformer, hsr-diff,Zhu_mamba_fusion,SSRMamba,Jiaxin,KSSANet}.
Bayesian inference methods have limited flexibility due to their dependence on data distribution and prior assumptions. Matrix decomposition methods disregard the three-dimensional structure of HSIs, making it challenging to capture complex spatial-spectral relationships. Tensor-based methods involve high computational costs, making large-scale data processing difficult. Furthermore, traditional methods exhibit limited generalization, restricting their adaptability to diverse datasets and scenarios.

In recent years, a large number of DL-based methods have emerged. Compared to traditional approaches, DL-based methods do not require extensive manual priors and can directly extract features from HSIs for modeling. SR models base on Convolutional Neural Networks (CNNs)~\cite{SRCNN, hsr-net} initially achieve remarkable results in various SR tasks. Subsequently, Transformer-based models~\cite{fusformer} make significant progress in HSR due to their powerful sequence modeling capabilities. Additionally, generative models have also been developed in parallel, with models base on Generative Adversarial Networks (GANs)~\cite{SRGAN, gan} widely applied in SR for their ability to generate more realistic and intricate high-frequency details. Research into Diffusion-based methods~\cite{hsr-diff, diff} is also advancing, particularly for their better detail generation and training stability compared to GAN models.

Despite the aforementioned deep learning methods integrating multilayer perceptrons (MLPs), CNNs, and Transformer architectures, which have introduced a series of innovative designs and significant performance improvements, their core modeling strategy is fundamentally still limited by the constraints of traditional linear modeling paradigms. Specifically, linear deep networks struggle to achieve an optimal balance between computational efficiency and image generation quality when dealing with high-dimensional spectral images~\cite{dimension}.

Recently, inspired by the Kolmogorov-Arnold representation theorem~\cite{kan_01, kan_02}, Kolmogorov-Arnold Networks (KANs)~\cite{KAN} have been introduced as an innovative computational core. By replacing linear weights at the network's edges with learnable univariate functions based on b-splines. By incorporating spline functions, neural networks are enabled to achieve a more fine-grained adjustment of spectral features with fewer parameters. Nonetheless, spline functions themselves struggle to escape the constraints of the \textit{Curse Of Dimensionality }(COD)~\cite{cod2, Cod1}. The groundbreaking contribution of HSR-KAN lies in its ingenious integration of KANs, MLPs and CNNs, creating a HSI-SR network that strikes an excellent balance between computational efficiency and the quality of image generation.

In this paper, we integrate KANs, MLPs, and CNNs to form a new neural network called HSR-KAN. Our starting point is to meticulously design an effective network structure that leverages strengths of each computational core. While circumventing COD, we strive for optimal balance between model computational efficiency and the quality of image generation. Taking into account fine-grained control ability of spline functions for sequence adjustment, we use three KAN layers in a series of linear transformations as fusion module (called KAN-Fusion) to efficiently fuse LR-HSI and HR-MSI. To avoid the potential COD that may arise from the simplistic stacking of KAN layers, we draw inspiration from SENet~\cite{SENET}, propose a new attention module called \textit{KAN Channel Attention Block }(KAN-CAB), which uses KAN layers only for adjusting the attention weight of spectral channels. KAN-CAB achieves precise spectral feature adjustment for post-fusion spectral feature with fewer parameters. In summary, HSR-KAN not only efficiently leverages the advantages of KAN but also cleverly avoids the problem of COD.

The main contributions of this work are summarized as follows:
\begin{itemize}  
    \item We propose a module for fusing LR-HSI and HR-MSI called KAN-Fusion.KAN-Fusion can finely fuse spectral sequence features with spatial texture features due to the introduction of KAN.
    
    \item Our proposed KAN-CAB module models the fused spectral features, and through the design of the channel attention structure, it exploits the advantages of KAN while avoiding the COD caused by the spline function.
    
    \item We propose an HSI-SR network called HSR-KAN. HSR-KAN mainly consists of two modules, KAN-Fusion and KAN-CAB, and HSR-KAN effectively integrates the advantages of different computing cores. Extensive experiments demonstrate that HSR-KAN achieves state-of-the-art results and achieves an outstanding balance between computational expense and SR quality. 
\end{itemize}  

\section{Related Works}
\subsection{Deep Learning-Based HSI-SR}
With the development of deep learning, deep neural networks have come to dominate the field of HSI-SR. These approaches treat the HSI-SR task as a nonlinear mapping from LR-HSIs to HR-HSIs, and utilizing gradient descent to fit the mapping network in order to generate optimal HR-HSIs. 

CNN is one of the most popular computational cores, is utilized by MHF-Net~\cite{MHF_Net} through deep unfolding techniques, creating a multi-layer interpretable super-resolution network based on CNN. SSRNet~\cite{SSR-Net} employs cross-model message insertion (CMMI) to integrate features from LR-HSI and high-resolution multispectral HR-HSI base on CNN. HSRnet~\cite{hsr-net} uses LR-HSI of the same scale as HR-MSI, extracting information through spatial attention mechanisms and channel attention mechanisms.
With the introduction of Transformer~\cite{Transformer}, which powerful attention modeling mechanism has been increasingly applied to HSI-SR tasks. As a pioneer, Fusformer~\cite{fusformer} achieves remarkable performance through a self-attention mechanism based on spectral channel features. DCTransformer~\cite{DCTransformer} captures interactions between different modalities using a directional paired multi-head cross-attention mechanism, efficiently modeling spectral images. With the development of diffusion generative models, diffusion methods have also been introduced into HSR tasks. HSR-Diff~\cite{hsr-diff} employs a conditional denoising mode to train the network, generating more Restructured spectral images. However, when dealing with high-dimensional spectral data, these methods often require very deep network structures or a large amount of training data to achieve satisfactory approximation results.
\subsection{Kolmogorov-Arnold Networks (KANs)}
Due to its fine-grained sequence modeling capability, KAN is initially applied to two-dimensional time series prediction. TKAN~\cite{TKAN} integrates KAN with LSTM networks for multi-step time series forecasting. However, this approach is not well-suited for effectively modeling 3D hyperspectral images (HSIs), as it fails to capture the spatial relationships between spatial pixels. U-KAN~\cite{U-KAN} incorporates KAN modules into the U-Net architecture for efficient medical image segmentation. However, for super-resolution tasks, U-KAN experiences pixel information loss after continuous encoding, and the decoding process is disrupted by interference from downsampling low-resolution layers. Further validation of the effectiveness of KAN in processing RGB images is provided in \cite{KAN_demo}. However, effectively leveraging KAN networks for modeling hyperspectral images remains a challenge.

\section{Proposed HSR-KAN Network}
\begin{figure*}[!htp]
    \centering
    \includegraphics[width=\linewidth]{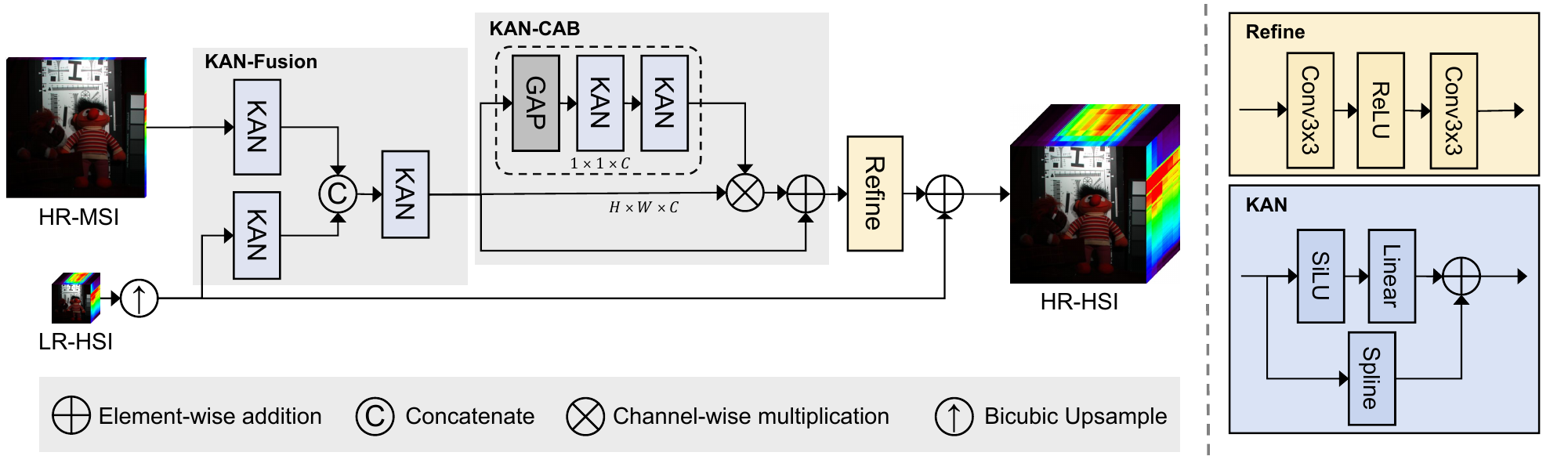}
    \caption{The structure diagram of HSR-KAN. "Conv" denotes a convolutional layer with a 3$\times$3 kernel, "ReLU" denotes the \textit{ReLU} activation function, "GAP" denotes the \textit{Global Average Pooling}, "Spline" denotes the \textit{B-spline function}, "SiLU" denotes \textit{SiLU} activation function. }
    \label{fig:overview}
\end{figure*}
\subsection{Problem Formulation}
The observation model for HR-MSI and LR-HSI is as follows:
\begin{equation}
    \mathbf{X}=\mathbf{RZ},\mathbf{Y}=\mathbf{ZD},
\end{equation}
 where $\mathbf{X} \in \mathbb{R}^{c \times H \times W}$ represents HR-MSI, $\mathbf{Y} \in \mathbb{R}^{C \times h \times w}$ represents LR-HSI, and $\mathbf{Z} \in \mathbb{R}^{C \times H \times W}$ represents HR-HSI.
Here, $W$ and $H$ denote the width and height of high-resolution image, respectively, while $w$ and $h$ denote the width and height of low-resolution image, respectively. $C$ and $c$ represent the number of spectral bands of hyperspectral and multispectral image, respectively.
$D \in \mathbb{R}^{HW \times hw}$ is the spatial response function of LR-HSI, representing the spatial degradation model of HR-HSI. $R \in \mathbb{R}^{b \times B}$ represents the spectral response function of HR-MSI, describing the spectral degradation model of HR-HSI.
HSI-SR can be defined as an inverse problem where a latent $\mathbf{Z}$ is generated from existing $\mathbf{X}$ and observed $\mathbf{Y}$. 

\subsection{Architecture Overview}
Fig.~\ref{fig:overview} illustrates the overall framework of HSR-KAN, which consists of three main parts: KAN-Fusion, multi-layer KAN-CAB, and the Restructure module. Initially, LR-HSI and HR-MSI are fused via KAN-Fusion to generate spectral latent features. These features are then input into the multi-layer KAN-CABs for feature extraction. Each layer of KAN-CAB consists of two KANs, forming a SENet~\cite{SENET} structure. Finally,  the Restructure module through two convolutional layers to adjust the generated latent features into the shape of HR-HSI.

\subsection{KAN Layer}
In KAN layer, each node is fully connected to every node in subsequent layer. For each edge, a separate, trainable activation function is applied. At each node, only a summation operation is performed over all incoming edges. As shown in the bottom right corner of Fig.~\ref{fig:overview}, learnable activation functions are defined as weighted sums of B-splines, denoted by the B-spline basis functions $B_i$, with the fixed residual function chosen as Sigmoid Linear Unit (SiLU):
\begin{equation}
    \text{KAN}(x)=w_1\cdot\text{SiLU}(x)+w_2\cdot\sum_{i=0}^{G+k-1}c_i\cdot B_i(x).
\end{equation}
The weights $w_1$, $w_2$ and the basis function coefficients $c_i$ are trainable parameters of spline. The basis function $B_i$ is chosen as a $ k $-th degree polynomial, with the default value $k = 3$. The grid parameter $G$ determines the degree of B-spline construction with the default value $ G = 5$.
In detail, for specified parameters $k$, $G$ and the domain $[t_0, t_G]$, a vectore of equidistant knot points $ \vec{t} = (t_{-k}, \ldots, t_0, \ldots, t_{G + k})$ is constructed. Subsequently, $G + k$ basis functions $ B_i^k(x)$ are defined recursively as follows:
\begin{equation}
    B_i^0(x)=\begin{cases}1&\text{if } t_i\leq x<t_{i+1},\\0&\text{otherwise},\end{cases}
\end{equation}
and for $k > 0$:
\small{
\begin{equation}
    B_{i}^{k}(x)=\frac{x-t_{i}}{t_{i+k}-t_{i}}B_{i}^{k-1}(x)+\frac{t_{i+k+1}-x}{t_{i+k+1}-t_{i+1}}B_{i+1}^{k-1}(x).
\end{equation}
}
By further allowing coefficient $c_i$ to locally adapt and alter the entire spline, KAN can fit arbitrary mapping functions without a specific form.
\subsection{KAN-Fusion}
Intuitively, spline functions allow for more direct adjustment of the importance weight of spectral sequence and spatial texture features, and the shallow design essentially avoids the COD. Based on this concept, we utilize a combination of triangular KAN layers for the fusion of LR-HSI and HR-MSI, with details as follows.

To facilitate the linear transformation by the KAN layer, we first fold the input HR-MSI $\mathbf{X}$ and the upsampled LR-HSI $UP(\mathbf{Y})$ in the spatial dimension:
\begin{align}
 \label{eq:X_reshape}   & \mathbf{X_0} = reshape(\mathbf{X}), \quad \mathbf{X_0} \in \mathbb{R}^{HW \times C}
    \\ 
  \label{eq:Y_reshape}  & \mathbf{Y_0} = reshape(UP(\mathbf{Y})), \quad \mathbf{Y_0} \in \mathbb{R}^{HW \times C} 
\end{align}
Next, as \eqref{eq:fus}$, \mathbf{X_0}$ and $\mathbf{Y_0}$ are individually mapped to a common dimension $D$ by a KAN layer, $D$ is set to 256 as a hyperparameter. Subsequently, a KAN layer aligns two concatenated vectors while maintaining the channel dimension as $D$. For ease of handling, we reshape the output $O_0 \in \mathbb{R}^{HW \times D}$ into a feature map of dimensions $O_0{'} \in \mathbb{R}^{D \times H \times W}$.
\begin{align}
    \label{eq:fus}& \mathbf{O_0} = KAN(concat[KAN(\mathbf{X_0}),KAN(\mathbf{Y_0})]) \\ 
    & \mathbf{O_0^{'}} = reshape(\mathbf{O_0}) \quad \mathbf{O_0} \in \mathbb{R}^{D \times H \times W}
\end{align}
Finally, the fused feature $\mathbf{O_0^{'}}$ is passed through the KAN-CAB module for feature extraction.
\subsection{KAN Channel Attention Block (KAN-CAB)}
HSR-KAN adopts $L$ consecutive KAN-CAB modules for spectral feature extraction. KAN-CAB module consists of an SENet structure composed of two KAN layers. Details are as follows. 

KAN-CAB can be mainly divided into two stages as Algorithm~\ref{code:KAN-CAB}. In first stage, \textit{Global Average Pooling} (GAP) is applied to compress the feature of each spectral channel in input feature map (e.g., for the feature map $ O_i$ of the $ i $-th layer). This means that for each spectral channel, we obtain a single value to represent its global importance. In the second stage, two consecutive KAN layers are utilized to learn importance weights of each channel. These weights are then used to generate a vector that reweights channels of the input feature map. Finally, we multiply the spectral channel weight vector with input $O_i$ channel-wise, precisely adjusting the global composition ratio of local spectral pixels. 

Employing a channel attention module integrated with KAN, rather than simply stacking KAN layers, effectively circumvents the COD. It is noteworthy that although the KAN layer incorporates an MLP, the initial intent of integrating the MLP is to concurrently learn the composite structure (extrinsic degrees of freedom) and univariate functions (intrinsic degrees of freedom). In essence, the KAN layer is proposed as an effective linear model. Corresponding to the HSI-SR task, the extraction of spatial structural information is equally crucial. Therefore, designing a module based on the spectral channel attention mechanism is a viable option. Such a module adeptly balances the extraction of spatial texture features with the modeling of spectral sequence information.
\begin{algorithm}[t]
\caption{PyTorch-style Implementation of KAN-CAB}
\label{code:KAN-CAB}
\begin{lstlisting}[style=pythonstyle]
class KAN_SA(nn.Module):
  def __init__(self,dim):
    self.dim = dim
    self.kan1 = KAN(dim,dim)
    self.kan2 = KAN(dim,dim)
    
  def forward(self, x):
    b,c,h,w = x.shape
    assert c == self.dim
    shorcut = x
    # b c h w -> b c 1 1
    score = F.adaptive_avg_pool2d(x,(1,1))
    score = score.reshape((b,c)) 
    score = self.kan1(score)
    score = self.kan2(score)
    score = score.reshape((b,c,1,1))
    # channel-wise multiplication
    x = x * score
    return x + shorcut
\end{lstlisting}
\end{algorithm}

\begin{table*}[!htb]
    \caption{Quantitative comparisons of different approaches were conducted on the CAVE, Harvard, and Chikusei test datasets. The best results are highlighted in \textbf{bold}, while the second-best results are \underline{underlined}.}
    \label{tab:CAVE}
    \setlength{\tabcolsep}{4pt} 
    \centering
    \begin{tabularx}{\linewidth}{l c*{12}{X}}
    \toprule
    \multirow{2}{*}{Methods} & 
    \multirow{2}{*}{Scale}   & 
    \multicolumn{4}{c}{CAVE} &  
    \multicolumn{4}{c}{Harvard} &  
    \multicolumn{4}{c}{Chikusei} \\
    \cmidrule(lr){3-6} \cmidrule(lr){7-10} \cmidrule(lr){11-14}
     & & PSNR$\uparrow$  & SSIM$\uparrow$ & SAM$\downarrow$ & ERGAS$\downarrow$ & 
         PSNR$\uparrow$ & SSIM$\uparrow$ & SAM$\downarrow$ & ERGAS$\downarrow$ & 
         PSNR$\uparrow$ & SSIM$\uparrow$ & SAM$\downarrow$ & ERGAS$\downarrow$ \\
    \midrule
         FUSE~\cite{Fuse}         &$\times$2&50.81&3.34&2.69&0.989& 51.04&2.31&2.28&0.992&  32.81&3.09&4.96&0.931 \\ 
         MHF-Net~\cite{MHF_Net}       &$\times$2&52.04&2.86&1.27&0.995& 43.20&4.02&12.11&0.981& 37.19&2.86&4.52&0.957 \\ 
         HSRnet~\cite{hsr-net}        &$\times$2&52.89&2.33&1.10&0.996& 50.29&2.43&2.34&0.989&  39.95&1.86&3.07&0.968 \\ 
         Fusformer~\cite{fusformer}     &$\times$2&53.03&2.26&1.07&0.996& 51.07&2.34&2.28&0.992&  41.26&1.69&2.77&0.970 \\ 
         DCTransformer~\cite{DCTransformer} &$\times$2&53.74&2.07&0.97&0.998& 51.61&2.29&2.24&0.994&  41.87&1.53&2.68&0.972 \\
         HSR-Diff~\cite{hsr-diff}      &$\times$2&\underline{53.98}&\underline{1.94}&\underline{0.90} &\underline{0.999}& \underline{51.68}&\underline{2.16}&\underline{2.13}&\underline{0.996}&  \underline{42.12}&\underline{1.48}&\underline{2.54}&\underline{0.978} \\ 
         HSR-KAN       &$\times$2&\textbf{54.62}&\textbf{1.86}&\textbf{0.79}&\textbf{0.999}& \textbf{51.81}&\textbf{2.10}&\textbf{1.86}&\textbf{0.997}&  \textbf{42.31}&\textbf{1.37}&\textbf{2.48}&\textbf{0.981} \\
    \midrule
         FUSE~\cite{Fuse}           &$\times$4&39.72&4.83&4.18&0.975& 42.06&3.23&3.14&0.977&  27.76&4.80&7.22&0.882 \\ 
         MHF-Net~\cite{MHF_Net}        &$\times$4&46.32&3.33&1.74&0.992& 40.37&4.64&24.17&0.966& 33.19&3.18&6.24&0.927 \\ 
         HSRnet~\cite{hsr-net}        &$\times$4&47.82&2.66&1.34&0.995& 44.29&2.66&2.45&0.984&  36.95&2.08&3.60&0.952 \\ 
         Fusformer~\cite{fusformer}      &$\times$4&48.56&2.52&1.30&0.995& 45.06&2.62&2.39&0.987&  37.01&2.04&3.54&0.958 \\ 
         DCTransformer~\cite{DCTransformer} &$\times$4&48.78&2.49&1.29&0.996& 46.34&2.59&2.36&0.989&  37.12&2.01&3.46&0.961 \\
         HSR-Diff~\cite{hsr-diff}      &$\times$4&\underline{48.86}&\underline{2.42}&\underline{1.27}&\underline{0.996}& \underline{46.52}&\underline{2.54}&\underline{2.32}&\underline{0.991}&  \underline{37.19}&\underline{1.98}&\underline{3.42}&\underline{0.964}\\ 
         HSR-KAN       &$\times$4&\textbf{49.17}&\textbf{2.40}&\textbf{1.25}&\textbf{0.997}& \textbf{47.12}&\textbf{2.48}&\textbf{2.29}&\textbf{0.993}&  \textbf{37.42}&\textbf{1.87}&\textbf{3.31}&\textbf{0.968 }\\
    \midrule
         FUSE~\cite{Fuse}          &$\times$8&36.24&8.64&6.49&0.818&      40.13&4.05&3.98&0.980&  26.81&6.21&10.04&0.869\\ 
         MHF-Net~\cite{MHF_Net}       &$\times$8&41.17&4.54&2.95&0.837&   42.16&3.99&30.17&0.916& 29.21&4.02&8.09&0.911 \\ 
         HSRnet~\cite{hsr-net}        &$\times$8&42.54&4.06&2.64&0.845&   41.29&4.08&4.85&0.961&  30.85&3.21&6.68&0.921 \\
         Fusformer~\cite{fusformer}     &$\times$8&43.21&3.89&2.41&0.869& 40.96&4.24&3.99&0.976&  33.96&2.86&4.45&0.939 \\ 
         DCTransformer~\cite{DCTransformer} &$\times$8&44.49&\underline{3.19}&\underline{2.08}&0.898& 41.86&3.89&3.15&0.980&  34.08&2.54&4.02&0.945 \\
         HSR-Diff~\cite{hsr-diff}      &$\times$8&\underline{44.54}&3.23&2.13&\underline{0.899}& \underline{41.91}&\underline{3.44}&\underline{3.12}&\underline{0.982}&  \underline{34.21}&\underline{2.31}&\underline{3.86}&\underline{0.949}\\ 
         HSR-KAN       &$\times$8&\textbf{44.89}&\textbf{3.17}&\textbf{2.04}&\textbf{0.991}& \textbf{42.14}&\textbf{3.40}&\textbf{3.09}&\textbf{0.984}&  \textbf{34.51}&\textbf{2.23}&\textbf{3.64}&\textbf{0.951} \\
    \midrule
         Best Value& -& $+\infty$&0&0&1& $+\infty$&0&0&1& $+\infty$&0&0&1 \\ 
    \bottomrule
    \vspace{-20pt}
    \end{tabularx}
\end{table*}
After extracting feature using $L$ layers of KAN-CAB, resulting in the output $ O_l $, it is then input into the Restructure mdoule to reshape it into the spatial shape of HR-HSI.
\subsection{Restructure Module and Loss Function}
Restructure module is used to restore the spatial shape of HR-HSI, the Restructure module is represented as follows:
\begin{equation}
    \mathbf{\bar{Z}} = \text{Conv}(\text{ReLU}(\text{Conv}(O_l))) + UP(\mathbf{Y}),
\end{equation}
here, $O_l$ denotes the output after feature extraction by $L$ layers of KAN-CAB. $\mathbf{\bar{Z}}$ represents the predicted HR-HSI. $\text{Conv}$ denotes a convolution operation with a kernel size of 3, while $\text{ReLU}$ denotes the activation function.

Intuitively, the introduction of spline functions indeed provides the model with a more refined adjustment capability. However, the risk of overfitting that may arise from fully connected KAN layers cannot be overlooked. Employing sparse spline functions is an effective strategy to address this challenge. It should be noted that what needs to be sparsed is not the weights, but the spline functions.
\begin{equation}
\ell_{\mathrm{total}}=|\mathbf{\bar{Z}} 
 - \mathbf{Z}|_{1}+\ell_{\mathrm{sparse}},
\end{equation}
Here, $\ell_{\mathrm{sparse}}$ represents the default activation sparsity loss in~\cite{KAN}.

We define the L1 norm of each spline activation function $\phi$ in HSR-KAN as the average magnitude across its $N_p$ inputs, given by:
\begin{equation}
    |\phi|_1 \equiv \frac{1}{N_p} \sum_{s=1}^{N_p} \left| \phi(x^{(s)}) \right|. 
\end{equation}

For a KAN layer $\Phi $ with $ n_{\mathrm{in}}$ inputs and $ n_{\mathrm{out}}$ outputs, the L1 norm of $ \Phi $ is defined as the sum of the L1 norms of all constituent activation functions:
\begin{equation}
    \left| \Phi \right|_1 \equiv \sum_{i=1}^{n_{\mathrm{in}}} \sum_{j=1}^{n_{\mathrm{out}}} \left| \phi_{i,j} \right|_1.
\end{equation}

Furthermore, the entropy of \( \Phi \) is defined as:
\begin{equation}
    S(\Phi) \equiv -\sum_{i=1}^{n_{\mathrm{in}}} \sum_{j=1}^{n_{\mathrm{out}}} \frac{\left| \phi_{i,j} \right|_{1}}{\left| \Phi \right|_{1}} \log \left( \frac{\left| \phi_{i,j} \right|_{1}}{\left| \Phi \right|_{1}} \right).
\end{equation}

The total training objective $\ell $ is composed of the prediction loss $|\mathbf{\bar{Z}} 
 - \mathbf{Z}|_{1} $, plus L1 and entropy regularization terms for all KAN layers, expressed as:
\begin{equation}
    \ell = |\mathbf{\bar{Z}} 
 - \mathbf{Z}|_{1} + \lambda \left( \mu_1 \sum_{l=0}^{L-1} \left| \Phi_l \right|_1 + \mu_2 \sum_{l=0}^{L-1} S(\Phi_l) \right),
\end{equation}

where $ \mu_1 $ and $ \mu_2 $ are coefficients that are typically set equal to 1, and $ \lambda $ is the hyperparameter that governs the overall strength of regularization is also set to 1 in HSR-KAN.

\section{Experiment}
\subsection{Datasets and Benchmark}
We use three public datasets for testing: CAVE~\cite{CAVE}, Harvard and Chikusei~\cite{Chikusei}. Specifically, the CAVE dataset consists of 32 scenes, each with a spatial dimension of 512$\times$512, and includes 31 spectral bands. We select 20 images for training and 11 images for testing. The Harvard database has 50 indoor and outdoor images captured under daylight conditions, along with 27 indoor images taken under artificial and mixed lighting. Each HSI in this database has a spatial size of 1392$\times$1040 pixels and contains 31 spectral bands. We crop the top-left part of each image (1000$\times$1000 pixels) and randomly select 10 images for testing. Chikusei is a remote sensing HSI with 2048$\times$2048 spatial pixels and 128 spectral channels. The upper-left corner of 1024$\times$2048 pixels is used for training data, while the remaining area is divided into 8 patches of 512$\times$512 pixels each for testing.

The comparative methods include the traditional method: FUSE~\cite{Fuse}. CNN-based methods include SSRNet~\cite{SSR-Net}, MHF-Net~\cite{MHF_Net}, and HSRnet~\cite{hsr-net}. Transformer-based methods include Fusformer~\cite{fusformer}, DCTransformer~\cite{DCTransformer}, and HSR-Diff~\cite{hsr-diff}, with HSR-Diff being based on diffusion methods. The selected image quality assessment metrics include Peak Signal-to-Noise Ratio (PSNR), Structural Similarity Index (SSIM), Spectral Angle Mapper (SAM), and Erreur Relative Globale Adimensionnelle de Synthèse (ERGAS), which comprehensively evaluate the fidelity, structural consistency, spectral accuracy, and overall reconstruction performance of the generated images.

\subsection{Implementation Details}
We implement our model using the PyTorch framework and conduct training on Nvidia A800 GPUs. The batch size is set to 32 to balance training stability and computational efficiency. The Adam optimizer is employed with an initial learning rate of $4 \times 10^{-4}$, which undergoes a decay by a factor of 0.1 every 100 epochs to ensure stable convergence. To achieve optimal performance, all models are trained for a total of 1000 epochs.

For consistency in comparison, we set the number of KAN-CAB layers ($L$) to 4. All comparison methods strictly adhere to the default settings provided in their respective papers or official open-source implementations to ensure fair benchmarking. The KAN layer is implemented using the \textit{efficient-kan} library\footnote{\url{https://github.com/Blealtan/efficient-kan}}, which is specifically designed for efficient Kernel Attention Networks. To ensure an objective evaluation of computational efficiency and model complexity, we use \textit{fvcore}\footnote{\url{https://github.com/facebookresearch/fvcore}} to measure key performance indicators, including the number of parameters and floating-point operations per second (FLOPs). 

Ground Truth (GT) patches are cropped to 64 $\times$ 64, and then downsampled to 16 $\times$ 16 to serve as LR-HSI patches.
The LR-HSI patches are generated from the GT patches by applying a Gaussian blur with a kernel size of 3$\times$3 and a standard deviation of 0.5. In addition, the spectral response function of the Nikon D700 camera~\cite{MSI} is used to generate HR-MSI patches. 
\subsection{Comparison with State of Arts Methods}
To comprehensively assess the effectiveness of HSR-KAN, we compare its performance against state-of-the-art hyperspectral image super-resolution methods on benchmark datasets. Our evaluation includes both quantitative and qualitative analyses, covering multiple scaling factors. In addition, we provide visual comparisons and error heatmaps to highlight the reconstruction quality achieved by different models.
\subsubsection{Results on CAVE.}
\begin{figure*}[!htp]
    \centering
    \vspace{-20pt}
    \includegraphics[width=1\linewidth]{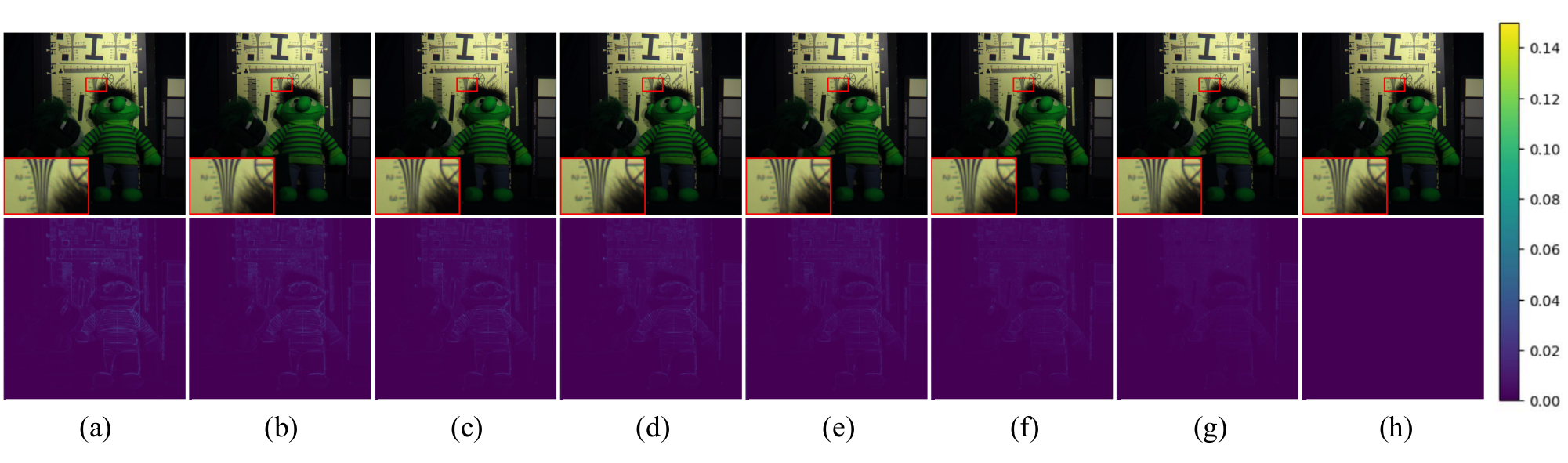}
    \vspace{-20pt}
    \caption{Visual quality comparison on the CAVE for $\times4$ SR, where the first row shows pseudo-color (R-20, G-30, B-2) images and second row shows corresponding heatmaps (mean squared error). (a) FUSE, (b) MHF-Net, (c) HSRnet, (d) Fusformer, (e) DCTransformer, (f) HSR-Diff, (g) HSR-KAN, (h) GT.}
    \label{fig:cave}
\end{figure*}

Table~\ref{tab:CAVE} presents the average image generation quality of 11 test images at scaling factors of $\times$2, $\times$4, and $\times$8. Among all evaluated models, HSR-KAN achieves the best performance across all quantitative metrics. Additionally, Fig.~\ref{fig:cave} showcases pseudo-colored images of charts and stuffed toys at a $\times$4 scaling factor, generated by different models, along with their corresponding Mean Squared Error (MSE) heatmaps. These visual comparisons demonstrate that all methods produce satisfactory results. However, HSR-KAN stands out with the lowest MSE, indicating its superior reconstruction accuracy compared to the GT.
\subsubsection{Results on Chikusei.}
\begin{figure*}[!htp]
    \centering
        \vspace{-10pt}
    \includegraphics[width=1\linewidth]{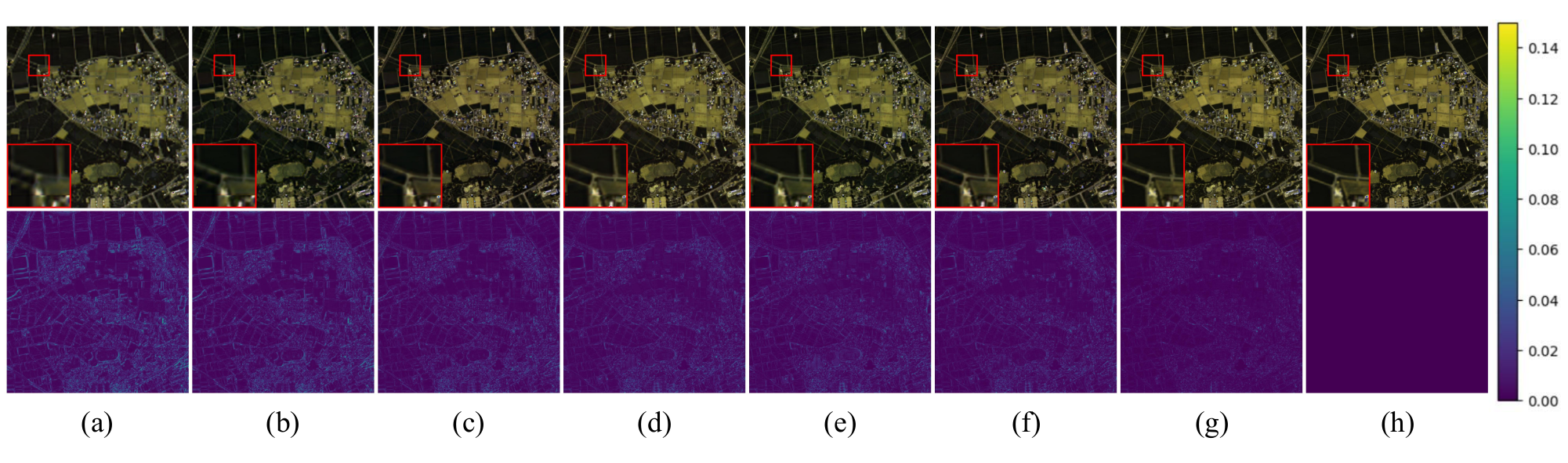}
    \vspace{-20pt}
    \caption{Visual quality comparison on Chikusei for $\times4$ SR, where first row shows pseudo-color (R-64, G-58, B-16) images and second row shows corresponding heatmaps. (a) FUSE, (b) MHF-Net, (c) HSRnet, (d) Fusformer, (e) DCTransformer, (f) HSR-Diff, (g) HSR-KAN, (h) GT.}
    \label{fig:chikusei}
\end{figure*}

Table \ref{tab:CAVE} illustrates the average image generation quality metrics for eight test patches from the Chikusei dataset at $\times2$, $\times4$, and $\times8$ scaling. HSR-KAN model exhibits significant enhancements across all evaluated metrics. Fig.~\ref{fig:chikusei} display pseudo-color images for $\times4$ SR generated by comparative models, accompanied by their respective MSE heatmaps. While all methods yield satisfactory visual outcomes, displayed heatmaps suggest that the image produced by our approach are more closely aligned with GT. These results show that HSR-KAN model is also effective to remote sensing HSIs.
\subsubsection{Generalization Performance on Harvard.} 
\begin{figure*}[!htp]
    \centering
        \vspace{-10pt}
    \includegraphics[width=1\linewidth]{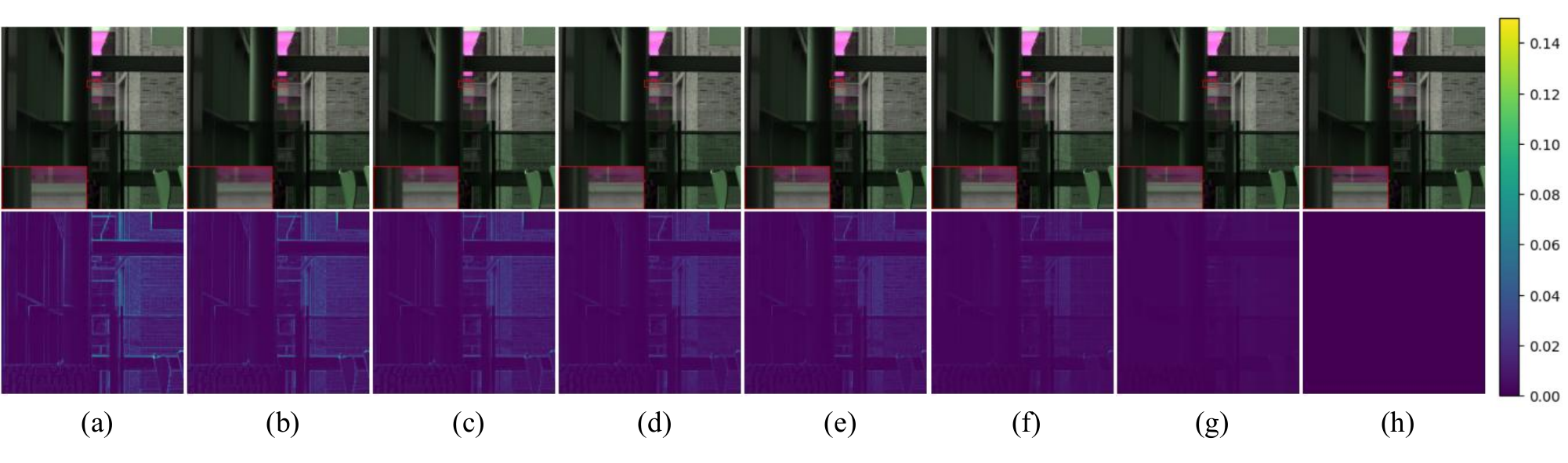}
    \vspace{-20pt}
    \caption{Visual quality comparison on Havard for $\times4$ SR, where first row shows the pseudo-color (R-29, G-22, B-31) images and second row shows corresponding heatmaps. (a) FUSE, (b) MHF-Net, (c) HSRnet, (d) Fusformer, (e) DCTransformer, (f) HSR-Diff, (g) HSR-KAN, (h) GT.}
    \label{fig:havard}
\end{figure*}

To validate the generalization performance of HSR-KAN, we test models trained on the CAVE dataset on the Harvard dataset \textit{without any additional training or fine-tuning}. We present the average image generation quality for all images in the Harvard dataset in Table \ref{fig:havard} of $\times2$, $\times4$ and $\times8$ scaling. Fig.~\ref{fig:havard} shows the pseudo-color super-resolution images for $\times4$ SR generated by various models on the Harvard dataset, along with their corresponding heatmaps. HSR-KAN demonstrates the best visual performance, without introducing additional noise or artifacts.  Quantitative experimental results and image generation results show that HSR-KAN achieves the best generalization performance.

In summary, HSR-KAN not only delivers outstanding performance in both quantitative and qualitative evaluations, consistently outperforming state-of-the-art methods across various metrics, but also demonstrates remarkable generalization capability. Its ability to effectively reconstruct high-quality hyperspectral images across different scaling factors highlights its robustness and adaptability. These results suggest that HSR-KAN is a highly effective solution for hyperspectral image super-resolution, balancing accuracy, efficiency, and generalizability.
\subsection{Ablation Study}
To gain deeper insights into the contributions of different components of our model, we perform a series of ablation experiments on the CAVE dataset for $\times$4 super-resolution. The performance of various ablation models is summarized in Table \ref{tab:ablation}, providing a comprehensive comparison of how each individual component impacts the overall performance. The detailed implementation of each ablation experiment is described below, highlighting the specific modifications made to isolate the effects of different model components.

\subsubsection{Impact of the KAN Layer}
 To comprehensively evaluate the significance of the KAN layer, we replace it with widely used computational modules, including MLP, CNN, and self-attention. The performance of these ablation models is presented in the first three rows of Table \ref{tab:ablation}. The results indicate that none of these alternative modules can surpass the original KAN-based design in terms of super-resolution accuracy. Specifically, MLP struggles to effectively capture complex spatial-spectral dependencies, CNN exhibits limitations in long-range feature modeling, and self-attention introduces high computational costs without yielding a corresponding improvement in performance. 

\subsubsection{Impact of KAN-Fusion} 
KAN-Fusion, a fusion module based on KAN, is compared with \textit{Stack Fusion}, where the LR-HSI is simply upsampled and concatenated with the HR-MSI, as utilized in Fusformer and HSR-Diff. The experimental results, presented in the fourth row of Table \ref{tab:ablation}, reveal a significant performance gap. Specifically, the PSNR of the Stack Fusion-based model is considerably lower than that of the KAN-Fusion-based baseline, highlighting the effectiveness of KAN-Fusion in enhancing image reconstruction quality.

\subsubsection{Impact of Sparse Loss}
The introduction of B-spline functions achieves finer granularity in weight adjustment, but it also tends to lead to overfitting. Therefore, the introduction of a regularized sparsity loss function is particularly necessary. To quantitatively analyze the effect of the sparsity loss function, we compare the performance of models with and without it, with the comparative results listed in the fifth row of Table \ref{tab:ablation}.
The variation in loss and PSNR values in Fig.~\ref{fig:ablation}. These exprimental results indicate that during the training process, initial convergence rate is faster without using a sparsity loss function, but as the model training progresses, the convergence rate significantly lags behind that of the training using a sparsity loss function. Overall, training HSR-KAN with a sparsity loss function can significantly enhance the upper limit of network performance.network performance.

\subsubsection{Impact of Channel Attention Block}
\begin{figure}
    \centering
    \includegraphics[width=1\linewidth]{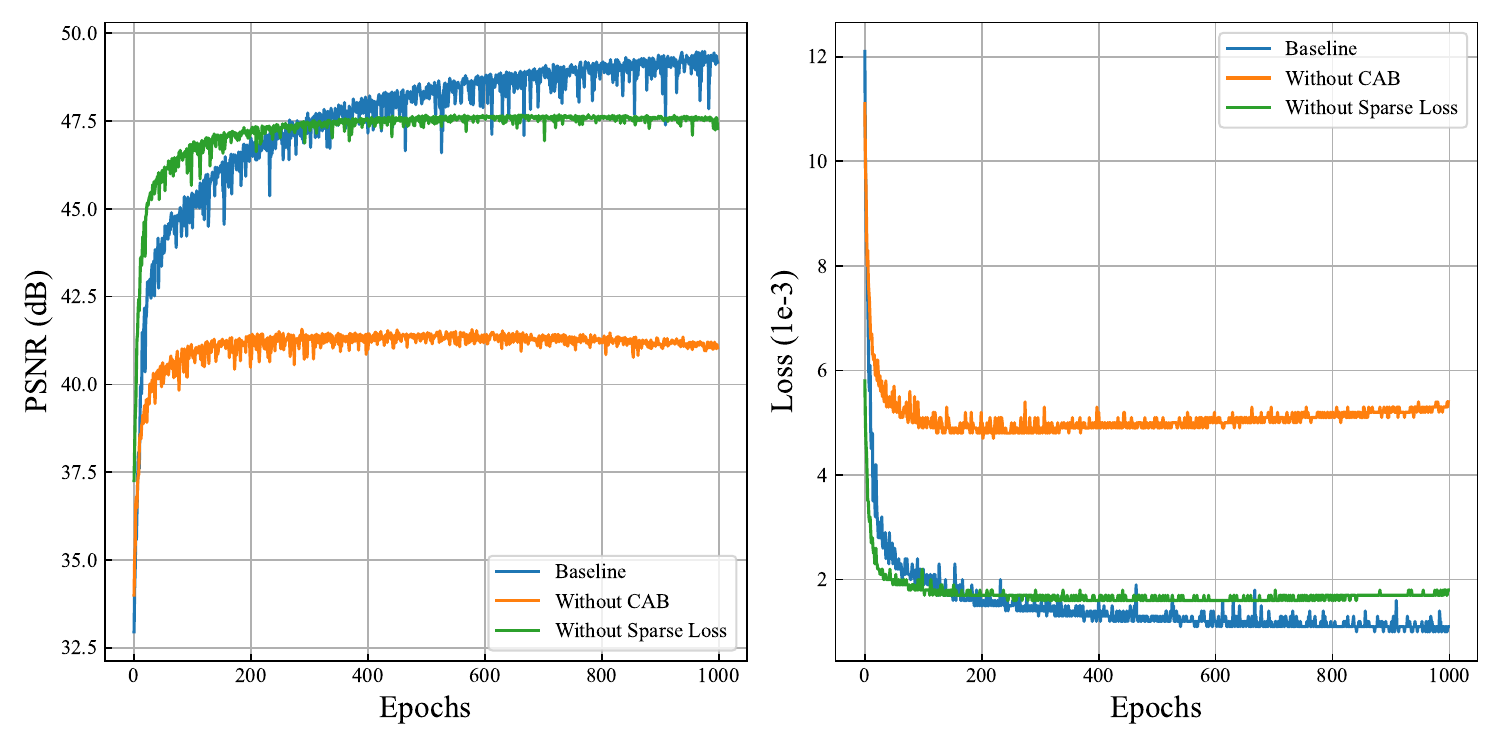}
    \vspace{-20pt}
    \caption{The full-size HSR-KAN model (Baseline), is compared with the two ablation models 'Without Sparse Loss' and 'Without CAB' for their training performance on the CAVE dataset for $\times4$ SR. The left graph plots the variation in PSNR values, while the right graph shows the changes in loss values.}
    \vspace{-10pt}
    \label{fig:ablation}
\end{figure}

Intuitively, integrating a channel attention mechanism with KAN can effectively alleviate COD. To quantitatively evaluate the impact of KAN-CAB, we compare it with a network that simply stacks KAN layers (denoted as "Without CAB" in Table \ref{tab:ablation}). Experimental results demonstrate that KAN-CAB not only reduces computational complexity but also significantly enhances performance. Furthermore, as shown in Fig.~\ref{fig:ablation}, it substantially accelerates network convergence. In summary, KAN-CAB effectively mitigates the impact of COD.
\begin{table}
\centering
\caption{Ablation quantitative results for HSR-KAN on CAVE datasets for $\times4$ SR. \#Params means the number of network parameters. \#FLOPs denotes the number of FLOPs.}
\label{tab:ablation}
\begin{tabularx}{\linewidth}{lXXXl}
\toprule
Ablation       & Variant                & \#Params(M) & \#FLOPs(G) & PSNR$\uparrow$ \\
\midrule
Baseline                              & -    & 7.30  & 7.87  & 49.17  \\ 
\midrule
\multirow{2}{*}{\centering Cores }   &  Conv & 6.78  &  27.19  & 41.34\\
                                      &  MLP  & 2.59  & 7.86  & 36.81  \\
                                      &  Self-Attention & 4.16 & 7.87  & 42.87    \\
\midrule
Fusion                                & Stack Fusion & 5.98 & 3.04 & 46.02  \\ 
Loss                                & without $\ell_{\mathrm{sparse}}$ & 7.30 & 7.87 & 47.28  \\ 
Attention                           & without \textit{CAB} & 7.30 & 27.19 & 41.03  \\ 
\bottomrule
\end{tabularx}
\end{table}

\begin{table}[!h]
    \centering
    \caption{Quantitative comparison results regarding different \textit{Spline Order} (with \textit{Grid Size} kept at 5). \#Params means the number of network parameters. \#FLOPs denotes the number of FLOPs.}
    \setlength{\tabcolsep}{8.5pt} 
    
    \resizebox{\linewidth}{!}{
\begin{tabular}{ccccc}
    \toprule
    Spline Order & \#Params(M) & \#FLOPs(G) & PSNR$\uparrow$ & SSIM$\uparrow$ \\
    \midrule
    1 & 6.25 & 7.87 & 39.12 & 0.975 \\
    3 & 7.30 & 7.87 & 49.17 & 0.997 \\
    5 & 8.35 & 7.87 & 49.20 & 0.997 \\
    7 & 9.40 & 7.87 & 49.23 & 0.997 \\
    9 & 10.45 & 7.87 & 48.18 & 0.996 \\
    \bottomrule
\end{tabular}
}
    \label{tab:spline_order}
\end{table}
\subsubsection{Impact of Spline Order}
To explore the impact of \textit{Spline Order} on network performance, we compare the network's performance under different \textit{Spline Order} settings (with \textit{Grid Size} kept at 5). From the results in Table \ref{tab:spline_order}, the \textit{Spline Order} increases, the size of the network model grows linearly while the FLOPs remain constant. However, when the \textit{Spline Order} exceeds 3, the SR performance of the network does not improve proportionally with the increase in model size. With a \textit{Spline Order} of 3, HSR-KAN achieves the optimal balance between SR performance and computational efficiency.

\subsubsection{Impact of Grid Size}
Different spline \textit{Grid Size} indicates the internal adjustment range of the learnable activation function. We further explore the impact of this parameter setting on the the SR performance of HSR-KAN (with \textit{Spline Order} kept at 3). As shown in Table \ref{tab:grid_size}, when the \textit{Grid Size} increases linearly, the model size also grows linearly, while the FLOPs remain unchanged. However, when the \textit{Grid Size} exceeds 5, there is no significant improvement in SR performance. Therefore, a \textit{Grid Size} of 5 represents the optimal balance point. 
\begin{table}[!h]
    \centering
    \caption{Quantitative comparison results regarding different \textit{Grid Size} (with \textit{Spline Order} kept at 3). \#Params means the number of network parameters. \#FLOPs denotes the number of FLOPs.}
    \setlength{\tabcolsep}{9.3pt} 
    \begin{tabularx}{\linewidth}{ccccc}
    \toprule
        Grid Size &  \#Params(M) & \#FLOPs(G) & PSNR$\uparrow$ & SSIM$\uparrow$ \\
    \midrule 
        1 & 5.21 & 7.87 & 36.28 & 0.812 \\
        3 & 6.25 & 7.87 & 40.28 & 0.843 \\
        5 & 7.30 & 7.87 & 49.17 & 0.997  \\
        7 & 8.35 & 7.87 & 49.02 & 0.997  \\
        9 & 9.40 & 7.87 & 48.71 & 0.996 \\
    \bottomrule
    \end{tabularx}
    \label{tab:grid_size}
    \vspace{-10pt}
\end{table}

\subsection{Comparison of Efficiency}
To assess the feasibility of model deployment, we conduct a preliminary evaluation of inference performance. The tests are performed on the CAVE dataset with a batch size of 1 for $\times$4 super-resolution. The experimental quantitative results are presented in Table \ref{tab:efficiency}. Compared to advanced super-resolution methods, such as HSR-Diff (based on diffusion) and DCTransformer (based on the Transformer architecture), our model achieves significant reductions in both model size and inference time. This is primarily due to the sparsity introduced by spline function computations and the lightweight design of the network architecture.

\begin{table}[!h]
    \centering
    \caption{Quantitative comparisons of computational performance on the CAVE dataset for $\times4$ SR. \#Params means the number of network parameters. \#FLOPs denotes the number of FLOPs.}
    \setlength{\tabcolsep}{6.8pt} 
    \begin{tabularx}{\linewidth}{lcccc}
    \toprule
        Method & \#Params(M) & \#FLOPs(G) & Time(ms)  & PSNR$\uparrow$  \\
    \midrule
        MFH-Net       &0.79  & 2.53  & 22.05  &46.32 \\ 
        HSRnet        &1.90  & 2.02  & 3.12   &47.82 \\ 
        Fusformer     &0.50  & 10.12 & 4.40   &48.56 \\ 
        DCTransformer &8.12  & 76.79 & 200.74 &48.78 \\ 
        HSR-Diff      &10.14 & 140.15& 16.14  &48.86 \\ 
        HSR-KAN       &7.30  & 7.87  & 6.18   &49.17 \\ 
    \bottomrule
    \end{tabularx}
    \label{tab:efficiency}
\end{table}

\section{Conclusion}
We propose an innovative hybrid neural network, HSR-KAN, for fusing a low-resolution hyperspectral image and a high-resolution multispectral image to generate a high-resolution hyperspectral image. HSR-KAN leverages the combined strengths of Kolmogorov-Arnold Networks, CNNs, and MLPs. Specifically, the integration of KANs enables finer-grained feature modeling, while the incorporation of MLPs and CNNs effectively mitigates the Curse of Dimensionality inherent in deep fully connected structures of KANs.
HSR-KAN achieves outstanding super-resolution performance while maintaining a compact model size and reduced computational complexity. Extensive experiments demonstrate that HSR-KAN consistently outperforms state-of-the-art methods, delivering superior reconstruction accuracy while ensuring high computational efficiency.
\section{Acknowledgments}
This research was supported by the National Key Research and Development Program of China (No. 2023YFB4502304).

\bibliography{main}

@misc{KAN,
      title={KAN: Kolmogorov-Arnold Networks}, 
      author={Ziming Liu and Yixuan Wang and Sachin Vaidya and Fabian Ruehle and James Halverson and Marin Soljačić and Thomas Y. Hou and Max Tegmark},
      year={2024},
      eprint={2404.19756},
      archivePrefix={arXiv},
      primaryClass={cs.LG},
      url={https://arxiv.org/abs/2404.19756}, 
}

@misc{U-KAN,
      title={U-KAN Makes Strong Backbone for Medical Image Segmentation and Generation}, 
      author={Chenxin Li and Xinyu Liu and Wuyang Li and Cheng Wang and Hengyu Liu and Yixuan Yuan},
      year={2024},
      eprint={2406.02918},
      archivePrefix={arXiv},
      primaryClass={eess.IV},
      url={https://arxiv.org/abs/2406.02918}, 
}

@misc{KAN_demo,
      title={Demonstrating the Efficacy of Kolmogorov-Arnold Networks in Vision Tasks}, 
      author={Minjong Cheon},
      year={2024},
      eprint={2406.14916},
      archivePrefix={arXiv},
      primaryClass={cs.CV},
      url={https://arxiv.org/abs/2406.14916}, 
}

@inproceedings{bayesian_01,  
 title={Bayesian sparse representation for hyperspectral image super resolution}, 
 url={http://dx.doi.org/10.1109/cvpr.2015.7298986}, 
 DOI={10.1109/cvpr.2015.7298986}, 
 booktitle={2015 IEEE Conference on Computer Vision and Pattern Recognition (CVPR)}, 
 author={Akhtar, Naveed and Shafait, Faisal and Mian, Ajmal}, 
 year={2015}, 
 month={Jun}, 
 language={en-US} 
 }

@article{bayesian_02,  
 title={Hyperspectral and Multispectral Image Fusion based on a Sparse Representation}, 
 journal={IEEE Transactions on Geoscience and Remote Sensing,IEEE Transactions on Geoscience and Remote Sensing}, 
 author={Wei, Qi and Bioucas-Dias, JoseM. and Dobigeon, Nicolas and Tourneret, Jean-Yves}, 
 year={2014}, 
 month={Sep}, 
 language={en-US} 
 }

@article{matrix_01,  
 title={Super-Resolution for Hyperspectral and Multispectral Image Fusion Accounting for Seasonal Spectral Variability}, 
 journal={Cornell University - arXiv,Cornell University - arXiv}, 
 author={Borsoi, RicardoAugusto and Imbiriba, Tales and Bermudez, J.C.M.}, 
 year={2018}, 
 month={Aug}, 
 language={en-US} 
 }

@article{Tensor_01,  
 title={Fusing Hyperspectral and Multispectral Images via Coupled Sparse Tensor Factorization}, 
 url={http://dx.doi.org/10.1109/tip.2018.2836307}, 
 DOI={10.1109/tip.2018.2836307}, 
 journal={IEEE Transactions on Image Processing}, 
 author={Li, Shutao and Dian, Renwei and Fang, Leyuan and Bioucas-Dias, Jose M.}, 
 year={2018}, 
 month={Aug}, 
 pages={4118–4130}, 
 language={en-US} 
 }

@article{fusformer,  
 title={Fusformer: A Transformer-based Fusion Approach for Hyperspectral Image Super-resolution}, 
 journal={Cornell University - arXiv,Cornell University - arXiv}, 
 author={Hu, Jin-Fan and Huang, Ting-Zhu and Deng, Liang-Jian}, 
 year={2021}, 
 month={Sep}, 
 language={en-US} 
 }

@article{gan,
  title={Generative adversarial networks},
  author={Goodfellow, Ian and Pouget-Abadie, Jean and Mirza, Mehdi and Xu, Bing and Warde-Farley, David and Ozair, Sherjil and Courville, Aaron and Bengio, Yoshua},
  journal={Communications of the ACM},
  volume={63},
  number={11},
  pages={139--144},
  year={2020},
  publisher={ACM New York, NY, USA}
}

@inproceedings{hsr-diff,
  title={HSR-Diff: Hyperspectral image super-resolution via conditional diffusion models},
  author={Wu, Chanyue and Wang, Dong and Bai, Yunpeng and Mao, Hanyu and Li, Ying and Shen, Qiang},
  booktitle={Proceedings of the IEEE/CVF International Conference on Computer Vision},
  pages={7083--7093},
  year={2023}
}

@article{diff,
  title={Denoising diffusion probabilistic models},
  author={Ho, Jonathan and Jain, Ajay and Abbeel, Pieter},
  journal={Advances in neural information processing systems},
  volume={33},
  pages={6840--6851},
  year={2020}
}

@article{hsr-net,  
 title={Hyperspectral Image Super-Resolution via Deep Spatiospectral Attention Convolutional Neural Networks}, 
 url={http://dx.doi.org/10.1109/tnnls.2021.3084682}, 
 DOI={10.1109/tnnls.2021.3084682}, 
 journal={IEEE Transactions on Neural Networks and Learning Systems}, 
 author={Hu, Jin-Fan and Huang, Ting-Zhu and Deng, Liang-Jian and Jiang, Tai-Xiang and Vivone, Gemine and Chanussot, Jocelyn}, 
 year={2022}, 
 month={Dec}, 
 pages={7251–7265}, 
 language={en-US} 
 }

@article{SRCNN,  
 title={Image Super-Resolution Using Deep Convolutional Networks}, 
 url={http://dx.doi.org/10.1109/tpami.2015.2439281}, 
 DOI={10.1109/tpami.2015.2439281}, 
 journal={IEEE Transactions on Pattern Analysis and Machine Intelligence}, 
 author={Dong, Chao and Loy, Chen Change and He, Kaiming and Tang, Xiaoou}, 
 year={2016}, 
 month={Feb}, 
 pages={295–307}, 
 language={en-US} 
 }

@inproceedings{SRGAN,  
 title={Photo-Realistic Single Image Super-Resolution Using a Generative Adversarial Network}, 
 url={http://dx.doi.org/10.1109/cvpr.2017.19}, 
 DOI={10.1109/cvpr.2017.19}, 
 booktitle={2017 IEEE Conference on Computer Vision and Pattern Recognition (CVPR)}, 
 author={Ledig, Christian and Theis, Lucas and Huszar, Ferenc and Caballero, Jose and Cunningham, Andrew and Acosta, Alejandro and Aitken, Andrew and Tejani, Alykhan and Totz, Johannes and Wang, Zehan and Shi, Wenzhe}, 
 year={2017}, 
 month={Jul}, 
 language={en-US} 
 }

@article{MHF_Net,  
 title={MHF-Net: An Interpretable Deep Network for Multispectral and Hyperspectral Image Fusion}, 
 url={http://dx.doi.org/10.1109/tpami.2020.3015691}, 
 DOI={10.1109/tpami.2020.3015691}, 
 journal={IEEE Transactions on Pattern Analysis and Machine Intelligence}, 
 author={Xie, Qi and Zhou, Minghao and Zhao, Qian and Xu, Zongben and Meng, Deyu}, 
 year={2022}, 
 month={Mar}, 
 pages={1457–1473}, 
 language={en-US} 
 }

@article{SSR-Net,  
 title={SSR-NET: Spatial–Spectral Reconstruction Network for Hyperspectral and Multispectral Image Fusion}, 
 url={http://dx.doi.org/10.1109/tgrs.2020.3018732}, 
 DOI={10.1109/tgrs.2020.3018732}, 
 journal={IEEE Transactions on Geoscience and Remote Sensing}, 
 author={Zhang, Xueting and Huang, Wei and Wang, Qi and Li, Xuelong}, 
 year={2021}, 
 month={Jul}, 
 pages={5953–5965}, 
 language={en-US} 
 }

@article{Transformer,  
 title={Attention is All you Need}, 
 journal={Neural Information Processing Systems,Neural Information Processing Systems}, 
 author={Vaswani, Ashish and Shazeer, Noam and Parmar, Niki and Uszkoreit, Jakob and Jones, Llion and Gomez, AidanN. and Kaiser, Lukasz and Polosukhin, Illia}, 
 year={2017}, 
 month={Jun}, 
 language={en-US} 
 }

@article{DCTransformer,
  title={Reciprocal transformer for hyperspectral and multispectral image fusion},
  author={Ma, Qing and Jiang, Junjun and Liu, Xianming and Ma, Jiayi},
  journal={Information Fusion},
  volume={104},
  pages={102148},
  year={2024},
  publisher={Elsevier}
}

@misc{TKAN,
      title={TKAN: Temporal Kolmogorov-Arnold Networks}, 
      author={Remi Genet and Hugo Inzirillo},
      year={2024},
      eprint={2405.07344},  
      archivePrefix={arXiv},
      primaryClass={cs.LG},
      url={https://arxiv.org/abs/2405.07344}, 
}

@misc{SENET,
      title={Squeeze-and-Excitation Networks}, 
      author={Jie Hu and Li Shen and Samuel Albanie and Gang Sun and Enhua Wu},
      year={2019},
      eprint={1709.01507},
      archivePrefix={arXiv},
      primaryClass={cs.CV},
      url={https://arxiv.org/abs/1709.01507}, 
}

@InProceedings{seg_01,
    author    = {Ren, Tianqi and Shen, Qiu and Fu, Ying and You, Shaodi},
    title     = {Point-Supervised Semantic Segmentation of Natural Scenes via Hyperspectral Imaging},
    booktitle = {Proceedings of the IEEE/CVF Conference on Computer Vision and Pattern Recognition (CVPR) Workshops},
    month     = {June},
    year      = {2024},
    pages     = {1357-1367}
}

@ARTICLE{Tracking_01,
  author={Islam, Mohammad Aminul and Xing, Wangzhi and Zhou, Jun and Gao, Yongsheng and Paliwal, Kuldip K.},
  journal={IEEE Transactions on Geoscience and Remote Sensing}, 
  title={Hy-Tracker: A Novel Framework for Enhancing Efficiency and Accuracy of Object Tracking in Hyperspectral Videos}, 
  year={2024},
  volume={62},
  number={},
  pages={1-14},
  doi={10.1109/TGRS.2024.3418337}}

@article{dimension,
  title={High-dimensional data analysis: The curses and blessings of dimensionality},
  author={Donoho, David L and others},
  journal={AMS math challenges lecture},
  volume={1},
  number={2000},
  pages={32},
  year={2000}
}

@inproceedings{Cod1,
  title={On the training of a Kolmogorov Network},
  author={K{\"o}ppen, Mario},
  booktitle={Artificial Neural Networks—ICANN 2002: International Conference Madrid, Spain, August 28--30, 2002 Proceedings 12},
  pages={474--479},
  year={2002},
  organization={Springer}
}

@article{cod2,  
 title={Space-filling curves and Kolmogorov superposition-based neural networks}, 
 url={http://dx.doi.org/10.1016/s0893-6080(01)00107-1}, 
 DOI={10.1016/s0893-6080(01)00107-1}, 
 journal={Neural Networks}, 
 author={Sprecher, David A. and Draghici, Sorin}, 
 year={2002}, 
 month={Jan}, 
 pages={57–67}, 
 language={en-US} 
 }

@inproceedings{kan_01,
  title={On the representation of continuous functions of many variables by superposition of continuous functions of one variable and addition},
  author={Kolmogorov, Andrei Nikolaevich},
  booktitle={Doklady Akademii Nauk},
  volume={114},
  number={5},
  pages={953--956},
  year={1957},
  organization={Russian Academy of Sciences}
}

@inbook{kan_02,  
 title={On the Representation of Continuous Functions of Several Variables as Superpositions of Continuous Functions of a Smaller Number of Variables}, 
 url={http://dx.doi.org/10.1007/978-94-011-3030-1_55}, 
 DOI={10.1007/978-94-011-3030-1_55}, 
 booktitle={Selected Works of A. N. Kolmogorov}, 
 author={Tikhomirov, V. M.}, 
 year={1991}, 
 month={Jan}, 
 pages={378–382}, 
 language={en-US} 
 }

@article{MSI,  
 title={Regularizing Hyperspectral and Multispectral Image Fusion by CNN Denoiser}, 
 url={http://dx.doi.org/10.1109/tnnls.2020.2980398}, 
 DOI={10.1109/tnnls.2020.2980398}, 
 journal={IEEE Transactions on Neural Networks and Learning Systems}, 
 author={Dian, Renwei and Li, Shutao and Kang, Xudong}, 
 year={2021}, 
 month={Mar}, 
 pages={1124–1135}, 
 language={en-US} 
 }

@article{Fuse,  
 title={Fast Fusion of Multi-Band Images Based on Solving a Sylvester Equation}, 
 url={http://dx.doi.org/10.1109/tip.2015.2458572}, 
 DOI={10.1109/tip.2015.2458572}, 
 journal={IEEE Transactions on Image Processing}, 
 author={ Qi Wei and Dobigeon, Nicolas and Tourneret, Jean-Yves}, 
 year={2015}, 
 month={Nov}, 
 pages={4109–4121}, 
 language={en-US} 
 }

@article{CAVE,  
 title={Generalized Assorted Pixel Camera: Postcapture Control of Resolution, Dynamic Range, and Spectrum}, 
 url={http://dx.doi.org/10.1109/tip.2010.2046811}, 
 DOI={10.1109/tip.2010.2046811}, 
 journal={IEEE Transactions on Image Processing}, 
 author={Yasuma, Fumihito and Mitsunaga, Tomoo and Iso, Daisuke and Nayar, Shree K}, 
 year={2010}, 
 month={Sep}, 
 pages={2241–2253}, 
 language={en-US} 
 }

@article{Chikusei,
  title={Airborne hyperspectral data over Chikusei},
  author={Yokoya, Naoto and Iwasaki, Akira},
  journal={Space Appl. Lab., Univ. Tokyo, Tokyo, Japan, Tech. Rep. SAL-2016-05-27},
  volume={5},
  number={5},
  pages={5},
  year={2016}
}

@article{Zhu_mamba_fusion,
  author  = {Zhu, Chunyu and Deng, Shangqi and Song, Xuan and Li, Yachao and Wang, Qi},
  journal = {IEEE Transactions on Geoscience and Remote Sensing},
  title   = {Mamba Collaborative Implicit Neural Representation for Hyperspectral and Multispectral Remote Sensing Image Fusion},
  year    = {2025},
  volume  = {63},
  number  = {},
  pages   = {1-15}
}

@inproceedings{KSSANet,
  author    = {Li, Baisong and Wang, Xingwang and Xu, Haixiao},
  booktitle = {ICASSP 2025 - 2025 IEEE International Conference on Acoustics, Speech and Signal Processing (ICASSP)},
  title     = {KSSANet: KAN-Driven Spatial-Spectral Attention Networks for Hyperspectral Image Super-Resolution},
  year      = {2025},
  pages     = {1-5},
  doi       = {10.1109/ICASSP49660.2025.10890177}
}

@inproceedings{SSRMamba,
  author    = {Li, Baisong and Wang, Xingwang and Xu, Haixiao},
  booktitle = {ICASSP 2025 - 2025 IEEE International Conference on Acoustics, Speech and Signal Processing (ICASSP)},
  title     = {SSRMamba: Efficient Visual State Space Model for Spectral Super-Resolution},
  year      = {2025},
  pages     = {1-5},
  doi       = {10.1109/ICASSP49660.2025.10887701}
}

@article{Jiaxin,
  author  = {Li, Jiaxin and Zheng, Ke and Yao, Jing and Gao, Lianru and Hong, Danfeng},
  journal = {IEEE Geoscience and Remote Sensing Letters},
  title   = {Deep Unsupervised Blind Hyperspectral and Multispectral Data Fusion},
  year    = {2022},
  volume  = {19},
  pages   = {1-5}
}

@article{class_02,
  author   = {Chen, Rong and Vivone, Gemine and Li, Guanghui and Dai, Chenglong and Hong, Danfeng and Chanussot, Jocelyn},
  journal  = {IEEE Transactions on Geoscience and Remote Sensing},
  title    = {Graph U-Net With Topology-Feature Awareness Pooling for Hyperspectral Image Classification},
  year     = {2025},
  volume   = {63},
  number   = {},
  pages    = {1-14},
  doi      = {10.1109/TGRS.2024.3511614}
}

@article{class_03,
  author   = {Ding, Shujie and Ruan, Xiaoli and Yang, Jing and Li, Chengjiang and Sun, Jie and Tang, Xianghong and Su, Zhidong},
  journal  = {IEEE Transactions on Geoscience and Remote Sensing},
  title    = {LRDTN: Spectral–Spatial Convolutional Fusion Long-Range Dependence Transformer Network for Hyperspectral Image Classification},
  year     = {2025},
  volume   = {63},
  number   = {},
  pages    = {1-21},
  doi      = {10.1109/TGRS.2024.3510625}
}

@article{seg_02,
  author   = {Xu, Aijun and Xue, Zhaohui and Li, Ziyu and Cheng, Shun and Su, Hongjun and Xia, Junshi},
  journal  = {IEEE Transactions on Geoscience and Remote Sensing},
  title    = {UM2Former: U-Shaped Multimixed Transformer Network for Large-Scale Hyperspectral Image Semantic Segmentation},
  year     = {2025},
  volume   = {63},
  number   = {},
  pages    = {1-21},
  doi      = {10.1109/TGRS.2025.3543821}
}

@article{OBJ_02,
  author   = {Xiong, Fengchao and Sun, Yongle and Zhou, Jun and Lu, Jianfeng and Qian, Yuntao},
  journal  = {IEEE Transactions on Geoscience and Remote Sensing},
  title    = {Spatial–Spectral–Temporal Correlation Filter for Hyperspectral Object Tracking},
  year     = {2025},
  volume   = {63},
  number   = {},
  pages    = {1-13},
  doi      = {10.1109/TGRS.2025.3546058}
}
\bibliographystyle{IEEEtran}
\clearpage
\end{document}